# Distributed Infrastructure Sensors and Cues for Robotic Fire-Fighting and Safety System on a Lunar Base

**Alexis Munguia[1], Eryc Rodriguez[2], Drake Russ[3], Jekan Thangavelautham[4], Ph.D.**

[1]Space and Terrestrial Robotic Exploration (SpaceTREx) Laboratory, University of Arizona, 1130 N Mountain Ave., N417, Tucson, AZ 85718; e-mail: alexisimt@arizona.edu,
[2]Space and Terrestrial Robotic Exploration (SpaceTREx) Laboratory, University of Arizona, 1130 N Mountain Ave., N417, Tucson, AZ 85718; e-mail: erycrod7@arizona.edu,
[3]Space and Terrestrial Robotic Exploration (SpaceTREx) Laboratory, University of Arizona, 1130 N Mountain Ave., N417, Tucson, AZ 85718; e-mail: drakeruss@arizona.edu,
[4]Space and Terrestrial Robotic Exploration (SpaceTREx) Laboratory, University of Arizona, 1130 N Mountain Ave., N417, Tucson, AZ 85718; e-mail: jekan@arizona.edu.

## ABSTRACT

There are renewed efforts to build a lunar base and house a team of astronauts on the Moon for extended periods. However, important challenges remain, such as a rapid response fire-hazard management system. In the event of a fire or toxic leak inside an early crewed lunar habitat, a single pressurized cylinder must be handled within minutes. However, suited astronauts and radio-delayed support from Earth cannot guarantee such response times. We describe a prototype fire-fighting and emergency response safety technology that embeds intelligence in the habitat itself: using thin ceramic QR codes affixed every few meters, each carries a local floor map, safe sensor limits, and step-by-step hazard responses. A low-power rover decodes a plate in a matter of seconds, polls the adjoining temperature-gas sensor cluster, and acts immediately, eliminating the need for a global map or large onboard database. In a representative layout, the interior is divided into roughly one code per two square meters. Distributing data this way reduces the required memory and significantly extends operations over a centralized map-centric design. Simulations show that rapid-response firefighting both simplifies the task and avoids costly resource use and indeterminate outcomes. Because knowledge is spread across passive plates, damage is localized, procedures can be updated by replacing a single code, and processor demands remain minimal.

## INTRODUCTION

There is renewed interest in returning humans to the Moon. Through the Artemis program, plans are afoot to setup human bases (Figure 1) that would enable astronauts to perform extended missions on the lunar surface. Longer-term habitation on the lunar surface presents some important challenges. These include mitigating emergency scenarios such as fire or toxic chemical leak. Such scenarios have led to attempts at rapid evacuation of the MIR and ISS stations. However, such strategies may not be sufficient for a lunar base due to its significant distance from Earth. Fire, in particular, can spread in a matter of minutes and incapacitate a team of human astronauts. Thus, safe shelters are needed to rapidly evacuate a human team of astronauts in-order to regroup and meaningfully respond to a fire.

We have been researching alternative solutions that utilize teams of autonomous firefighting robots that can perform rapid fire response. Fire-fighting robots can respond immediately and continue to handle a fire even under extreme scenarios such as being surrounded by fire and smoke. Such a solution is credible thanks to advancements in sensors, particularly IR cameras that can be used to track and respond to fires during the early stages of ignition.

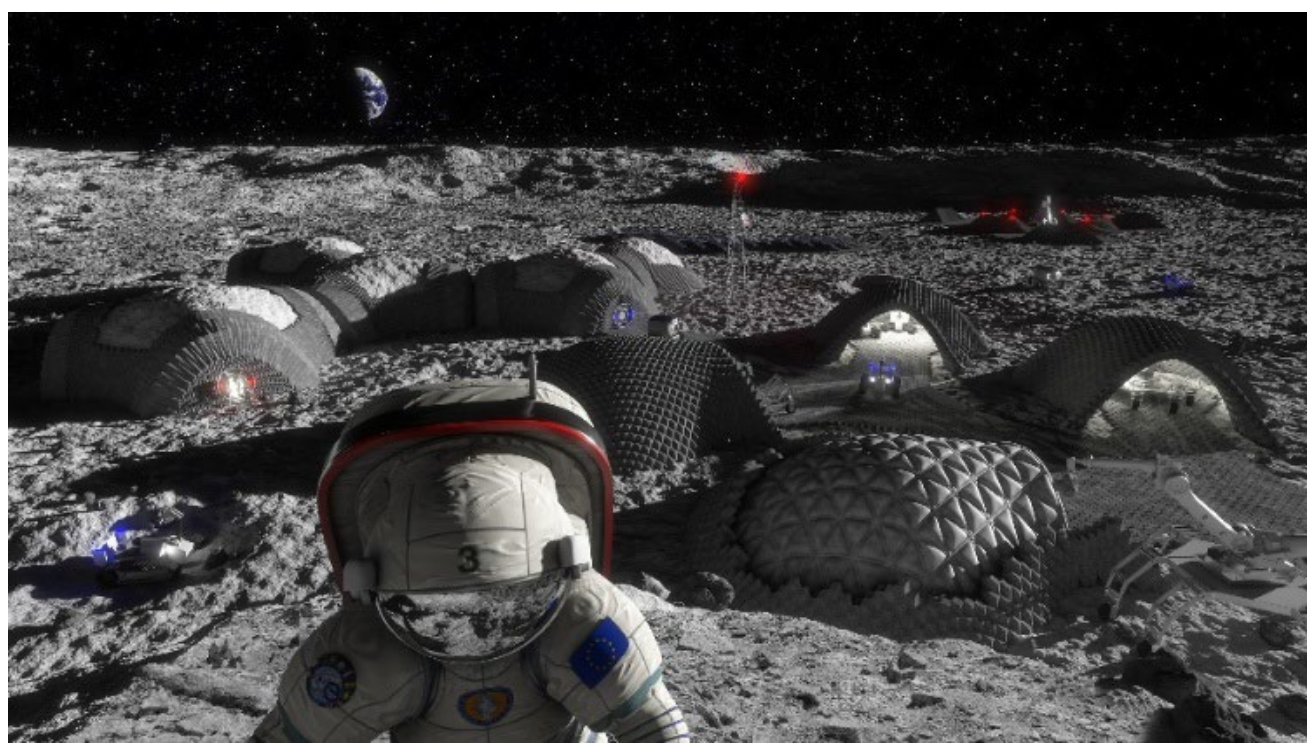

**Figure 1: ESA's 2018 Concept for a Lunar Base.**

In this work we develop a novel rapid-response system consisting of fire-fighting robots and sensor networks and cues that exploit use of fire-spread models to make autonomous tactical decisions to minimize the overall cost and damage from a fire threat. The technology builds upon our Tile Computing Network framework (Qui et al., 2023) but generalizes it further using multi-functional QR code-based visual cues that are used to rapidly partition, map/locate, assess fire-threats, obtain specialized knowledge about fire hazards, material combustibility, and chemical energy present and then finally develop a fire-fighting response plan that seeks to minimize overall costs.

The approach, as we show, permits rapid response but is not entirely reactive; rather, it plans a firefighting strategy that minimizes overall costs, damage, and risk. For such an assessment to be credible, the firefighting robot needs to have an updated assessment of the combustible materials, their amount, and their location within the room, particularly relative to the fire. The permissible action may include evacuating areas with fire hazards first, forming fire barricades, and implementing other creative strategies to prevent a fire from spreading. Further, the system models fire evolution using simplified cellular automata that can run on nimble edge-computing platforms and permit the evaluation of many variations, including detailed sensitivity analysis. Importantly, through the evaluation of fire-spread models, the system can gain insight into big-picture threats and work towards mitigating them.

This science-centric approach to fire-fighting stands in contrast to current fighting techniques that are largely reactive, in which overwhelming amount of water resources is used to first control the fire and then choke it. Such an option is simply not practical for a lunar base given the general lack of resources, including water. There is often insufficient granularity to avoid combustible fire-threats that could prolong efforts to limit fire-spread or utilization of resources. When the fire does spread beyond a control cordon, more fire-fighting resources are drawn in reactively.

The approach is novel in several respects, first enabling the geometrical partitioning of rooms and spaces to apply divide-and-conquer problem-solving techniques using QR-code cues. Next, the strategy permits the approach to be used even when the room/space is partially damaged or destroyed or is dynamically changing. This is an important distinction from conventional simulation-centric planning models. Such traditional models can lose correspondence with reality (the simulation correspondence/drift problem) due to complex, chaotic events, thereby losing their predictive utility. In contrast, our use of QR-code cues better enforces correspondence between a virtual model (digital twin) and reality, thereby keeping the fire-spread model accurate and relevant. Finally, QR codes and cues can withstand high temperatures and partial damage, and remain detectable even under smoke occlusion.

Beyond firefighting, technology may have other uses, such as utilizing swarms or multiple robot systems in resource-constrained, complex, dynamic, and chaotic environments. In the following section, we present related work, the overall system concept, analysis, and preliminary simulations, and conclude with a brief discussion.

## RELATED WORK

Embedding information into the environment for robot navigation is a well-established concept in industrial automation, where speed and precision are critical. Most notably, some warehouse robots navigate by reading markers embedded into the environment, such as QR Codes or bar codes, placed on the warehouse floor, walls, or ceiling. They are the cornerstone of supply management and logistics (McCathie, 2003). The code-based system is used to keep track of inventory and for product packaging/assembly. Bar-codes originated nearly 70 years ago and started as one-dimensional. The provided basis for simple data storage has limited storage and organizational capacity. More recently, two-dimensional barcodes have been introduced that better utilize surface area to store data.

Barcodes come in two categories: one-dimensional (1D or linear) and two-dimensional (2D). Linear barcodes encode data with lines of varying widths and spacing but have limited data storage capacity. To overcome this issue, 2D barcodes were introduced (Szentandrási et al., 2012). Their structure allows data to be stored on both vertical and horizontal axes, offering greater capacity compared to 1D barcodes. Further enhancements have been made to incorporate color 2D-codes that offer even greater data storage capacity (Taveerad and Vongpradhip, 2015). Further these QR-codes have been widely used for localization, navigation and data-encoding in the environment. (Sneha et al., 2020), (Kuo and Lin, 2025). Using embedded QR-Codes as visual cues, we are focused on simplifying off-world firefighting and emergency response. Putting out fire in space and aerospace vehicles is a critical step. The Apollo 1 disaster had highlighted what could go wrong (Guibaud et al., 2022).

Further, fire safety in extraterrestrial habitats such as lunar surface present unique challenges due to the reduced gravity fundamentally alters combustion behavior. As a result, flames tend to be enveloped into slow growing spheres around their fuel (Friedman, 1990). In the absence of convection, heat transfer to the surrounding atmosphere is greatly reduced, therefore continuously increasing the temperature of objects under thermal stress. One of the great challenges expected of lunar firefighting is the unique dynamics of fire spread under microgravity conditions. Fires in such circumstances can propagate with a lower concentration of oxygen and with a different velocity profile than under normal gravity (Urban et al., 2019).

Lunar gravity in specific has been described as a "goldilocks zone” that can prolong the presence of flames and delay self-extinguishing (Johnson and Ferkul, 2025), (Zayac et al., 2024).

These changes in flame behavior highlight the challenges for flame suppression aboard lunar stations and predicts the need for specialized firefighting strategies tailored for low-gravity. Our previous work has analyzed use of autonomous robots to mitigate and put out fires on a lunar base (Muniyasamy et al., 2024).

**Centralized vs Distributed Intelligence**

Given the off-world environment and low-gravity conditions, use of simulations and digital twins environments are of paramount importance to both analyze and predict flame spread behavior in these unique conditions. Digital twins are virtual representations of a larger model relying on a constant influx of data to enable monitoring and predictive control of a system. Digital twin concepts were invented in the aerospace industry to predict and simulate the failure of the Apollo 13 oxygen tank (Allen, 2021). Digital twins can provide simulations for scenarios like fire spreads or robot movements, which can prove valuable in stable, structured environments such as a factory or base. In unpredictable disasters, however, their current preferred configuration, namely server-centralized data-heavy reliance, can prove challenging if not catastrophic.

In a fire scenario on a lunar base, sensors or communications could be quickly damaged and disrupted by heat, rapid loss of base sustaining resources such as power, possibly rendering the centralized model insufficient. Furthermore, the reliance on heavy computational loads to produce predictions taxes computational resources that are limited and under stress during a time of emergency. In a rapidly escalating fire emergency, seconds of computation (compounded with the error introduced by downed sensors, simulation correspondence/drift challenges) could mean the difference between successful containment or a catastrophe. To summarize, centralized systems, with limited sensing and limited resources such as power may become unreliable when the environment changes at a pace faster than a digital model can update. These expected challenges suggest an alternative digital-twin architecture is needed.

Use of distributed intelligence, deriving popularity from simple biological organisms (Brook, 1999) offers a compelling solution in this scenario, where multiple semi-autonomous agents and information sources interact without a single, unified means of contact/communication. The independent, collaborative efforts of this system highly promote the odds of successful containment of hazards, even upon the steady loss of multiple nodes or agents. Given the independent nature of the system, the global outcome is met regardless of the failure of a single node or sector. By encoding mission-critical information into the environment itself, any singular agent is able to respond and take action when needed. In such a system, there is no single computing system that can crash. In emergency robotics, a distributed system means individual agents don't rely on constant validation or communication with a central server. Robots have higher individual agency, allowing them to discover instructions based on the current state of its surroundings and sensor-node networks. Distributed designs also simplify the integration of a network of simultaneously acting devices and sensors (Lopes et al., 2025).

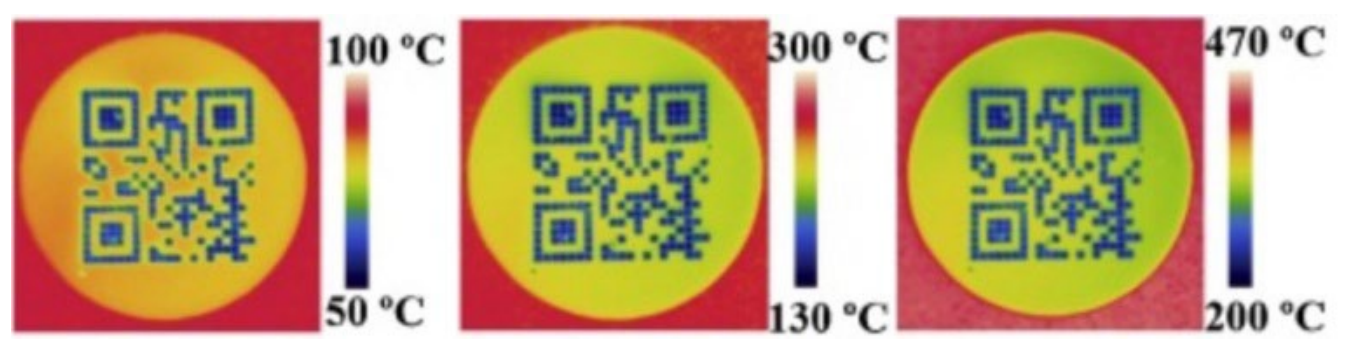

**Figure 2. High-temperature resistant code test using ceramic coating. Image reproduced from Fang et al., Variable optical properties of $La_{0.5}Sr_{0.5}Co_{1-x}Ni_xO_{3-\delta}$ for high-temperature resistant code application (2020).**

Perhaps the most challenging task when embedding information into fire-prone environments is the risk of losing valuable information. Visual indicators may be occulted by thick smoke and wireless communication devices may be damaged or destroyed. Heat-prone, heat-visible QR codes (Figure 2) offer a compelling alternative (Fang et al., 2020). Where a centralized architecture may be greatly hindered at the loss of information, a distributed intelligence system may be able to turn the perceived loss into a useful insight. The loss of a point of intelligence in a hazardous scenario may give insight into the current state of the environment.

## SYSTEM CONCEPT

The essence of our system design relies on a network of sensor nodes. These sensor nodes are composed of passive, data-dense, high-temperature robust QR Code tiles (Figure 3) affixed to the habitat/base (see Figure 4). These durable tags are strategically placed within the environment to divide the interior into a grid of localized information zones (Figure 4, right). Each QR code is printed on a piece of extreme-temperature-resistant material, such as ceramic-coated stainless steel, to withstand disaster situations. Industrial ceramic-coated QR codes have been shown to withstand extreme temperatures and be identifiable at temperatures as hot as 500°C (Fang et al., 2020). The QR-Code Tiles in turn form part of a large, distributed Tile Network (Figure 5) together with robots and sensor networks to perform base/facility wide tasks (Qiu et al., 2023).

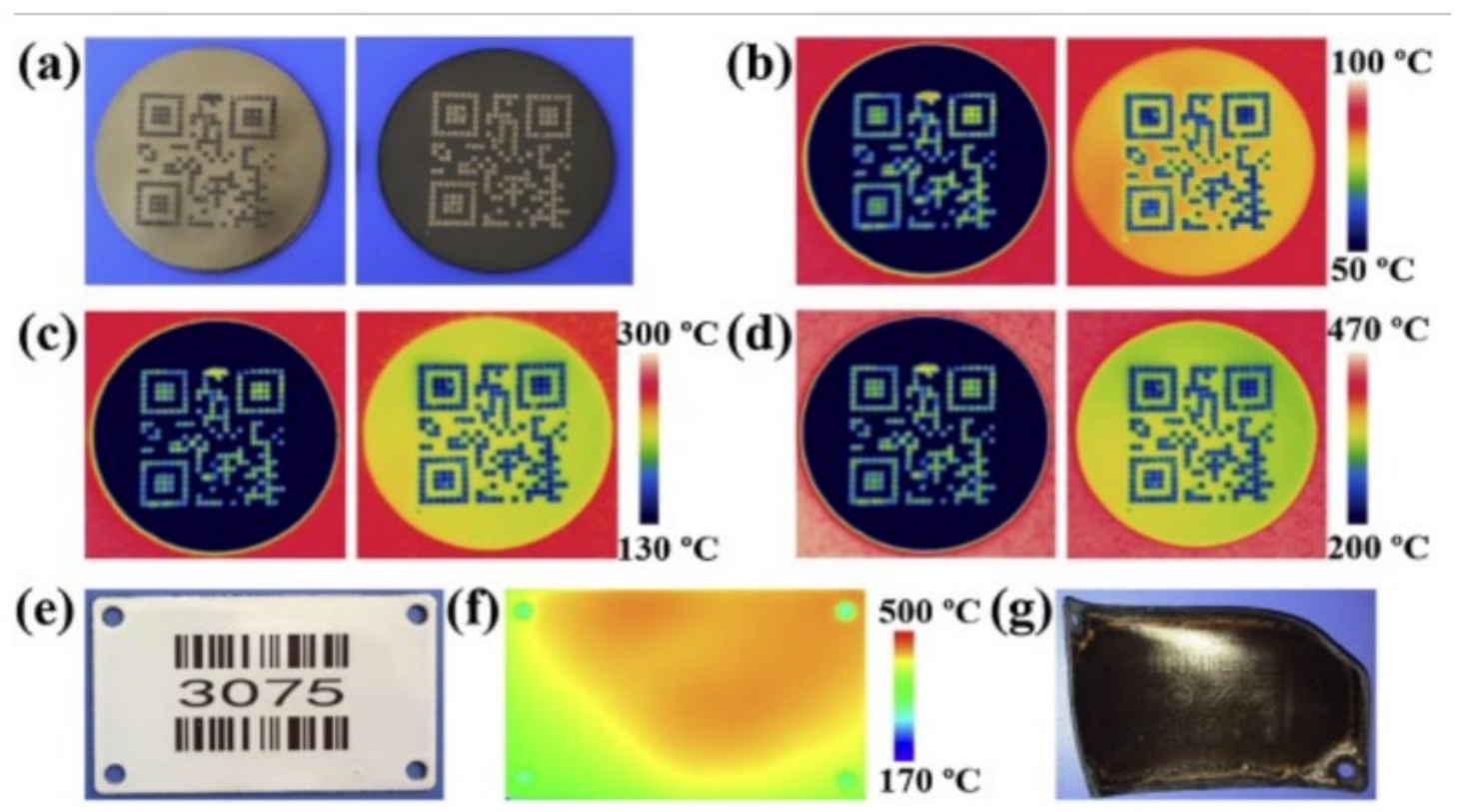

**Figure 3: Various examples of QR-code cues that can withstand high temperatures.**

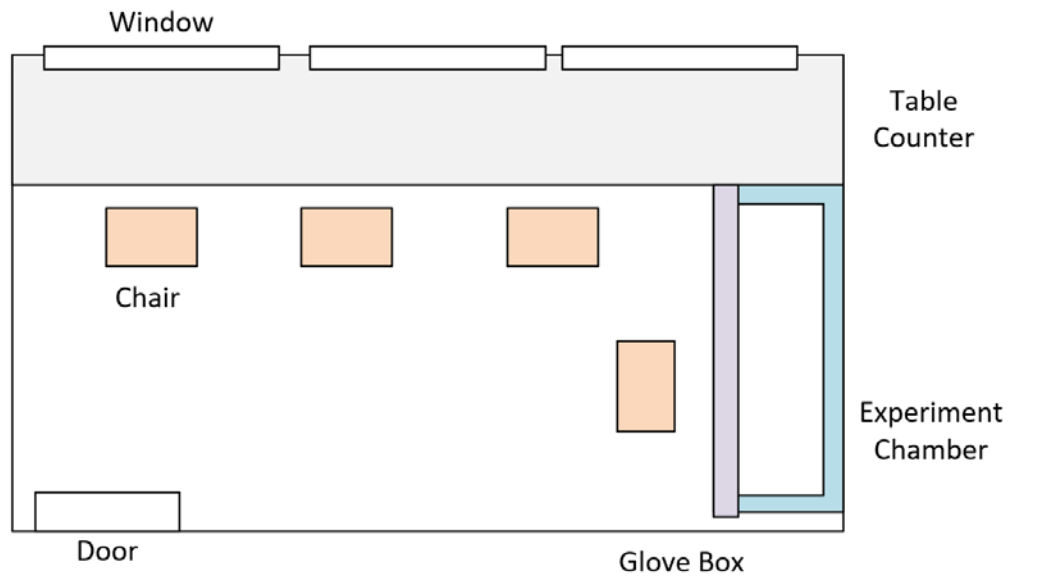

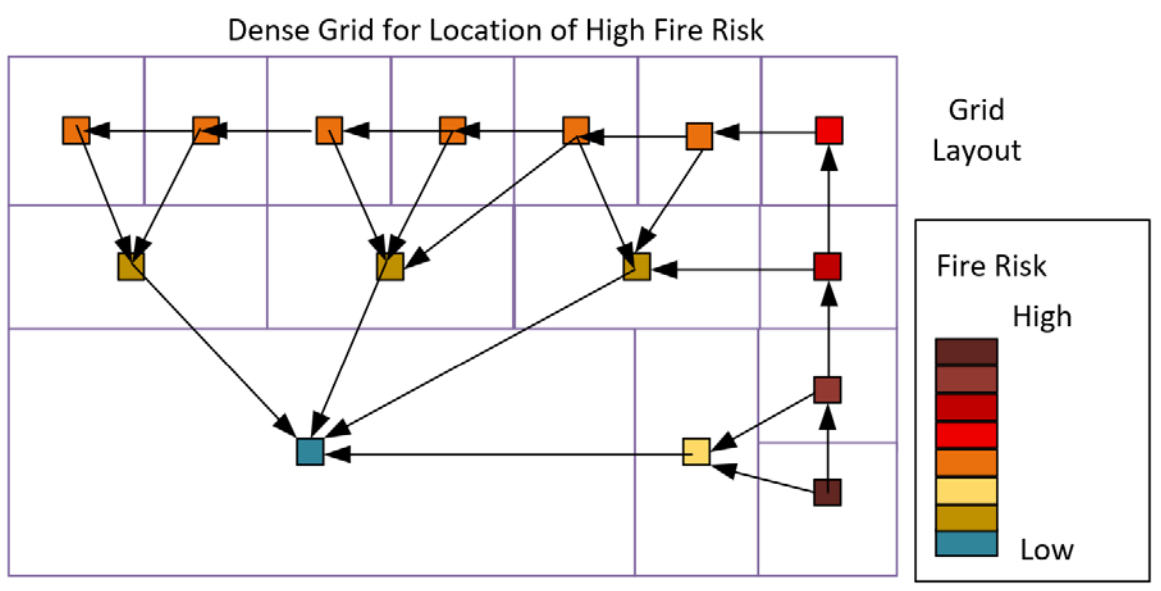

**Figure 4: (Left) Geochemistry laboratory on a lunar base. (Right) Grid layout of the laboratory for fire-hazard management. Small squares show the location of our QR-Code Tiles (QCT). Embedded within them is a fire-risk assessment of their sense-grid region. Further, QCTs provide a robot with information to scan and determine likely fire paths using arrowed lines.**

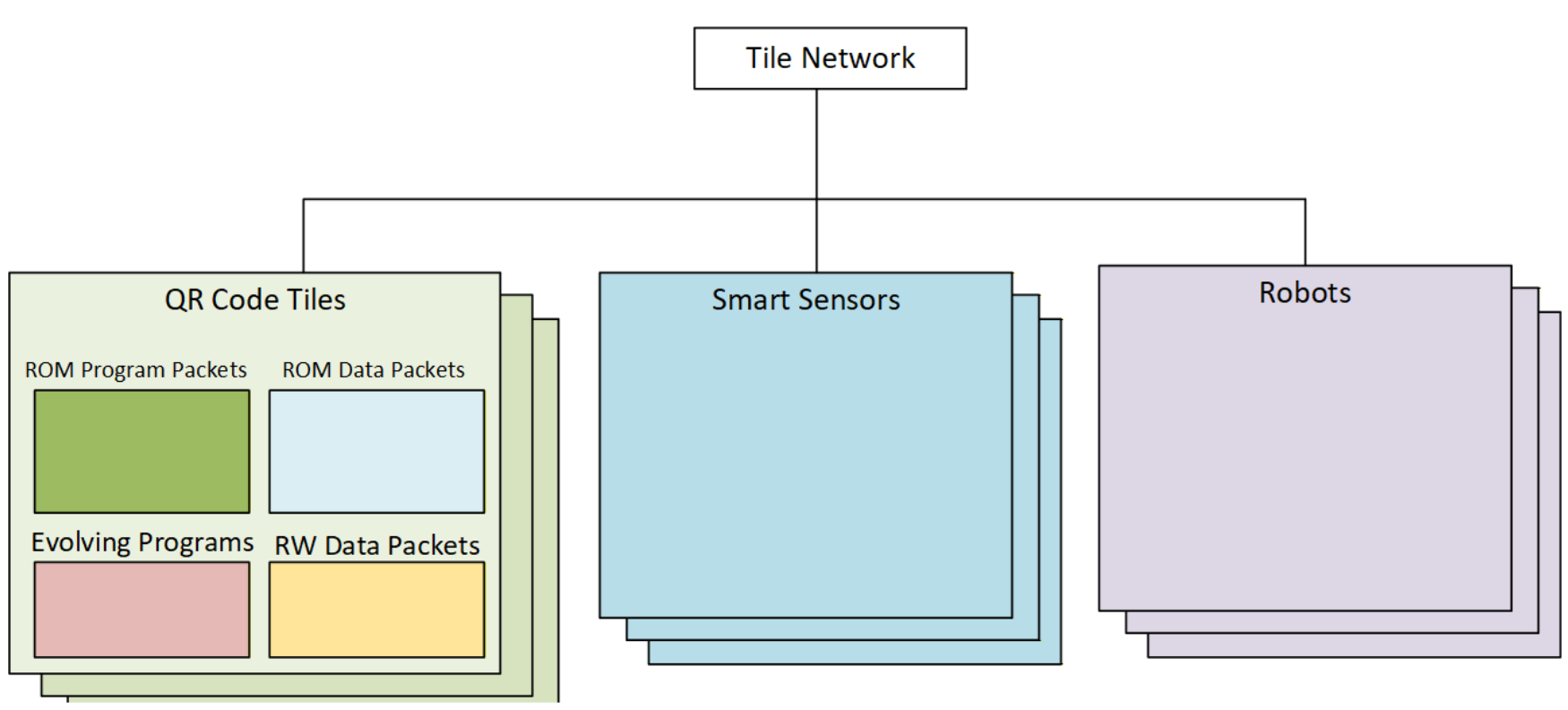

**Figure 5: The QR-Code Tiles extends the existing Tile Network architecture that promotes a distributed ecosystem of smart sensors, robots, and instruments working to robustly solve base/facility-wide tasks and do so by spatially distributing and offloading the knowledge and intelligence needed (Qiu et al., 2023).**

Metal or ceramic QR codes are exceptionally resilient, withstanding dangers from both extreme temperatures, rough handling, and chemical exposure with minimal degradation. This material property allows our system to ensure that QR code encoded information remains legible to robots, even in the event of an intense fire. Being that the tags act passively, that is without need for active electronics, the system itself is remarkably robust. This property's benefit is especially compounded in a lunar setting where reliability during disasters is paramount.

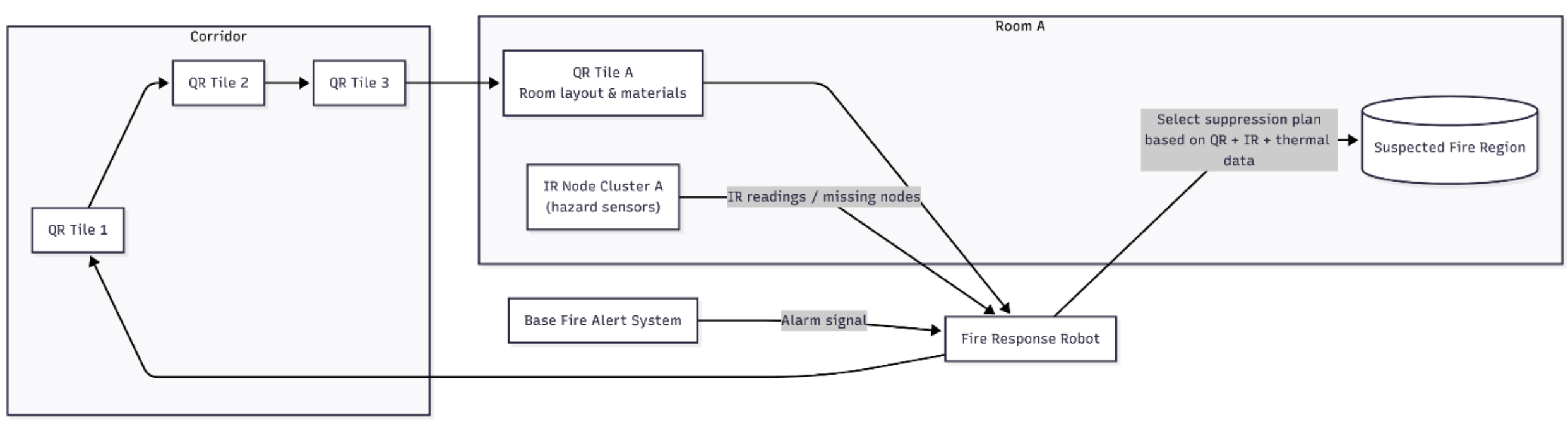

**Figure 6: QCTs refer to one another within the network. This enables the packing of invaluable multi-disciplinary information relevant to firefighting.**

Each QR Code Tile encodes valuable information pertaining to the environment's layout, materials, and emergency scenario strategies (Figure 6). By encoding a local floor plan segment into a habitat/room/lab, useful spatial information and fire-hazard threats can be delegated to robots rather quickly. A local floor map might include the immediate layout near the tag (room shape, entrances, exits, nearby storage of materials, inventory of hazardous materials and flammability). By leveraging the use of this readable information, robots may be able to orient themselves and maneuver around a space without the need for a global map. Another efficient use of spatial information encoding is the tracking of known possible dangers. The QCTs based on Figure 5 will embed visual information using QR codes, while other info, including evolving program and RW data, may be stored using an embedded RF chip and rewritable display technology.

By tracking information such as the potentially hazardous contents of a habitat's inventory, quick assessments may be made about the likely point of fire-ignition origin in the event of a disaster situation. Lastly, by knowing the contents of a room beforehand, specific step by step instructions may be laid out to be executed in the event of a disaster in that specific location. For example, a room containing materials prone to a chemical fire may encode specific instructions for optimal extinguishing of that specific fire as different fires require different methods of suppression. By encoding tailored instructions onto the QR Code Tiles (QCTs), the system effectively delegates scenario specific knowledge to the places where it is relevant.

If a room contains sensitive electronics, robots may be instructed to use $CO_2$ extinguishers. This granular distribution of instructions means the robot agents have no need for large onboard databases of all possible procedures; it simply reads instructions applicable to where it happens to be at that moment. The memory burden on the robot is therefore mitigated as the environment itself becomes the data repository for managing emergencies. By implementing such a system, the probabilistic nature of interpretations of on-board sensor readings is replaced with specific instructions and a higher degree of certainty. To account for losses and damage to QCTs, additional QCTs maybe placed and backup kept at the most fire-resistant location in the room.

**Sensor Nodes**

In addition to the QR Code Tiles (QCTs) based environmental intelligence, an extension of the system involves affixing infrared emitters to potential fire starting materials or high risk equipment throughout the lunar habitat. These emitters would broadcast a low power IR signal that can be detected and localized by the robots using its onboard sensors. Such a system would provide a reliable, nonspatial cue for locating potential fire sources or flammable material in vision impairing environments and serve as an implicit environmental gauge, whose state of operation may provide insight into the evolving state of a fire.

One of the primary challenges in fire fighting settings is the threat of degrading optical conditions due to thick smoke or heat distortion. The QR codes in our system, although robust and resistant to extreme environments, ultimately rely on visual readability. By introducing a system of IR nodes, the system is less susceptible to visual obstructions. Robotic agents are able to passively detect IR signals through thick smoke or darkness and triangulate the origin, enabling them to identify and navigate towards high priority targets when visual information is impaired. Further, we are evaluating embedded RF technology to provide an alternate data flow in case visual or IR options are compromised.

Each IR emitter would be encoded with a unique signal ID corresponding to the specific material or equipment it is fixed to (lithium-ion batteries, chemical solvents, oxygen lines), allowing the robot to easily identify a room's hazards and their spatial orientation. Combined with the robots' thermal sensing unit, the robot may easily track a heat signature and associate that with a potential source of the fire as well as be able to detect further dangers as the fire spreads. This allows both for a higher margin of confidence in detecting the origin of fires as well as better selection of fire suppression methods based on origin type.

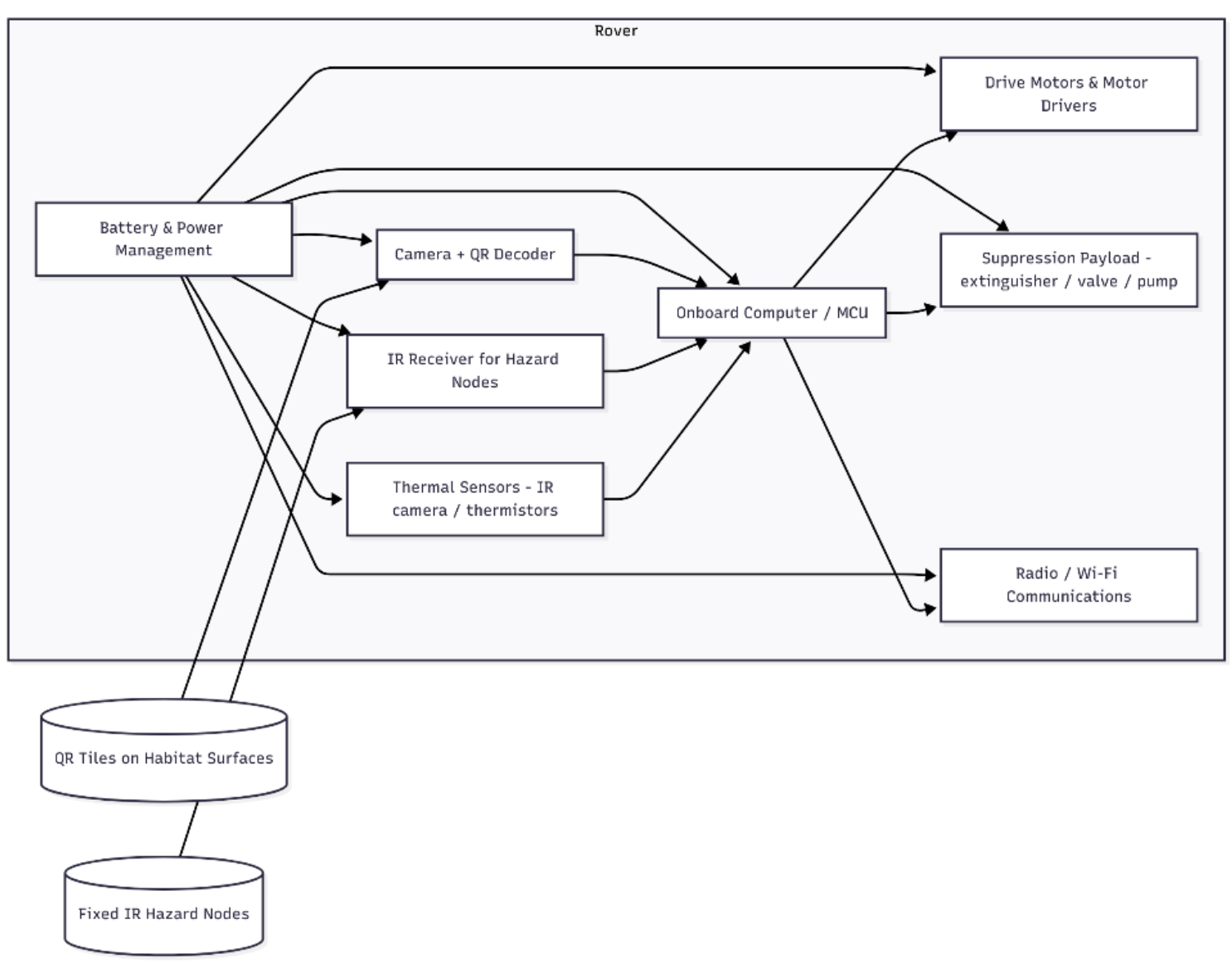

**Figure 7: Overall architecture of robot(s) and their subsystems interacting with QCTs.**

**Sensor Loss as a Diagnostic Insight**

A novel property of this design is that the loss of an IR emitter may itself be informative. In centralized systems, the failure of a sensor is typically seen as a loss of ability in detecting surroundings. However, in our proposed distributed intelligence system, the absence of an expected IR emitter may be reinterpreted as an indication of thermal compromise. If an IR emitter previously mapped to a hazardous material is no longer detectable, the robot can infer that the fire has likely destroyed the emitter and its associated hazard. This in conjunction with onboard thermal sensor data may provide agents with a strong indication as to the current conditions of a fire spread. This approach exemplifies how distributed intelligence systems can leverage gaps in data as indicators of changing conditions.

**Implications on Robot hardware**

To detect hazards and operate in potentially degraded conditions, each robot agent will be fixed with a sensor suite. The onboard sensors will include optical cameras, a thermal imaging sensor, and infrared receivers. The optical cameras will be mainly used for QR code scanning, acting as a means to receive information encoded throughout the habitat (see Figure 7). Cameras may be fitted with LED's or IR illuminators in the event of low light conditions. Thermal imaging will allow the robot to detect heat signatures and triangulate potential fire sources. A great advantage of this comes in the form of increasing redundancy, allowing robots to 'visualize' its surroundings, even in hazy conditions where visibility is low. Infrared beacon receivers provide a second non-visual cue to habitat conditions. By referencing current habitat IR emitter signals with information received via QR codes, inferences may be made about potential sources and direction of a fire.

Given the conditions, robot movement will likely require a novel solution. One such approach would be to use a ceiling-mounted rail system running along the length of each pressurized module (Figure 8). Robots may traverse the rails using motorized wheels, essentially acting as a suspended monorail. This traversal option minimizes obstruction by crew and materials in otherwise crowded conditions. It also enables rapid, unhindered movement throughout each of the base's compartments, where a traditional wheeled rover would be bottlenecked by floor obstacles and the computation required for navigation. This approach combines the stability of a fixed track with the flexibility of autonomous control.

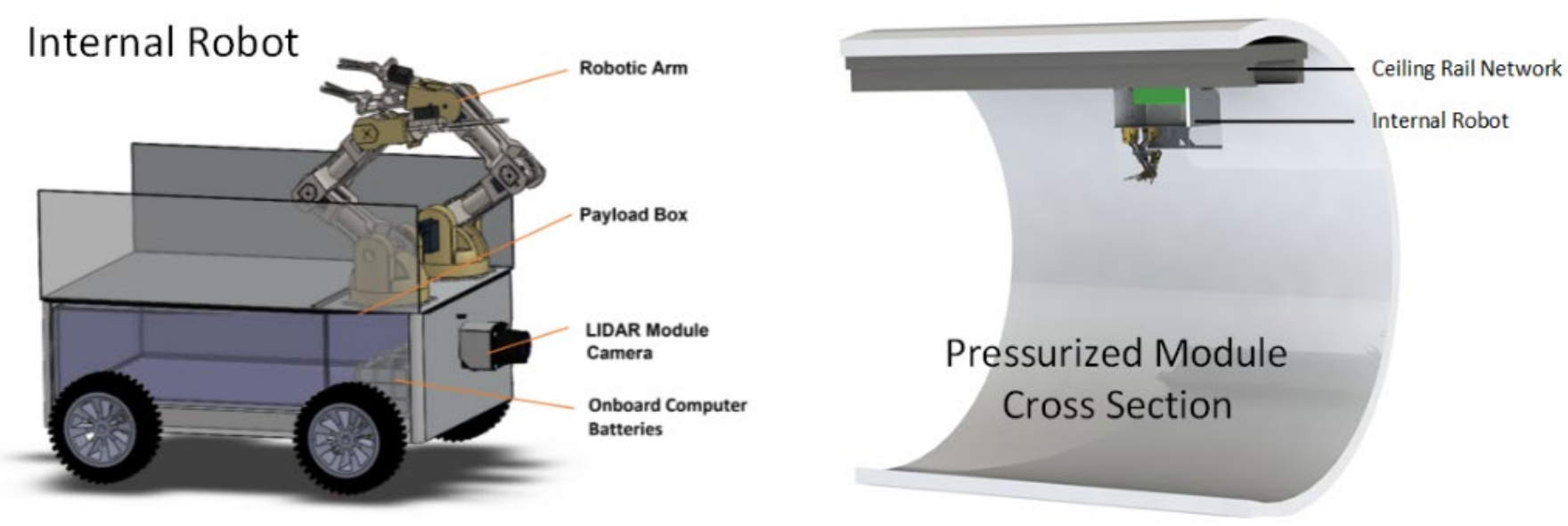

**Figure 8: Internal robot Concept (Left). The conceptual diagram of an internal robot moving on the ceiling of a pressurized modulus (Right) (Muniyasamy et al., 2024).**

Robot computational hardware is intentionally kept light, instead relying on a distributed tile network for coordination. This approach avoids resource-intensive computations and deductions in favor of direct information relaying. All detailed information of hazards is conveyed directly to a robot agent immediately upon its arrival to a partition of the habitat via a QR code. Robots will constantly interact only with information streams essential to them at their specific location and given task.

**Figure 9: Fire-fighting Behavior Architecture**

## ROBOT BEHAVIOR AND CONTROL ARCHITECTURE

One or more robots will use the QCTs to perform fire mitigation, following the overall behavior architecture (Figure 9). The feedback loop consists of detection and navigation, followed by overall assessment using model evaluation/prediction, and then actions, namely the evacuation of items and objects, particularly high-value or highly flammable substances, and a parallel effort to suppress, control, or extinguish the fire. The info and fire-fighting strategies are adapted for the local conditions. This increases the chance of fire-fighting success and further accounts for the unique characteristics of the fire-fighting scenario. This approach reduces surprises and reduces misalignment with on-the-ground reality. Further details pertaining to each step is outlined below (Figure 10):

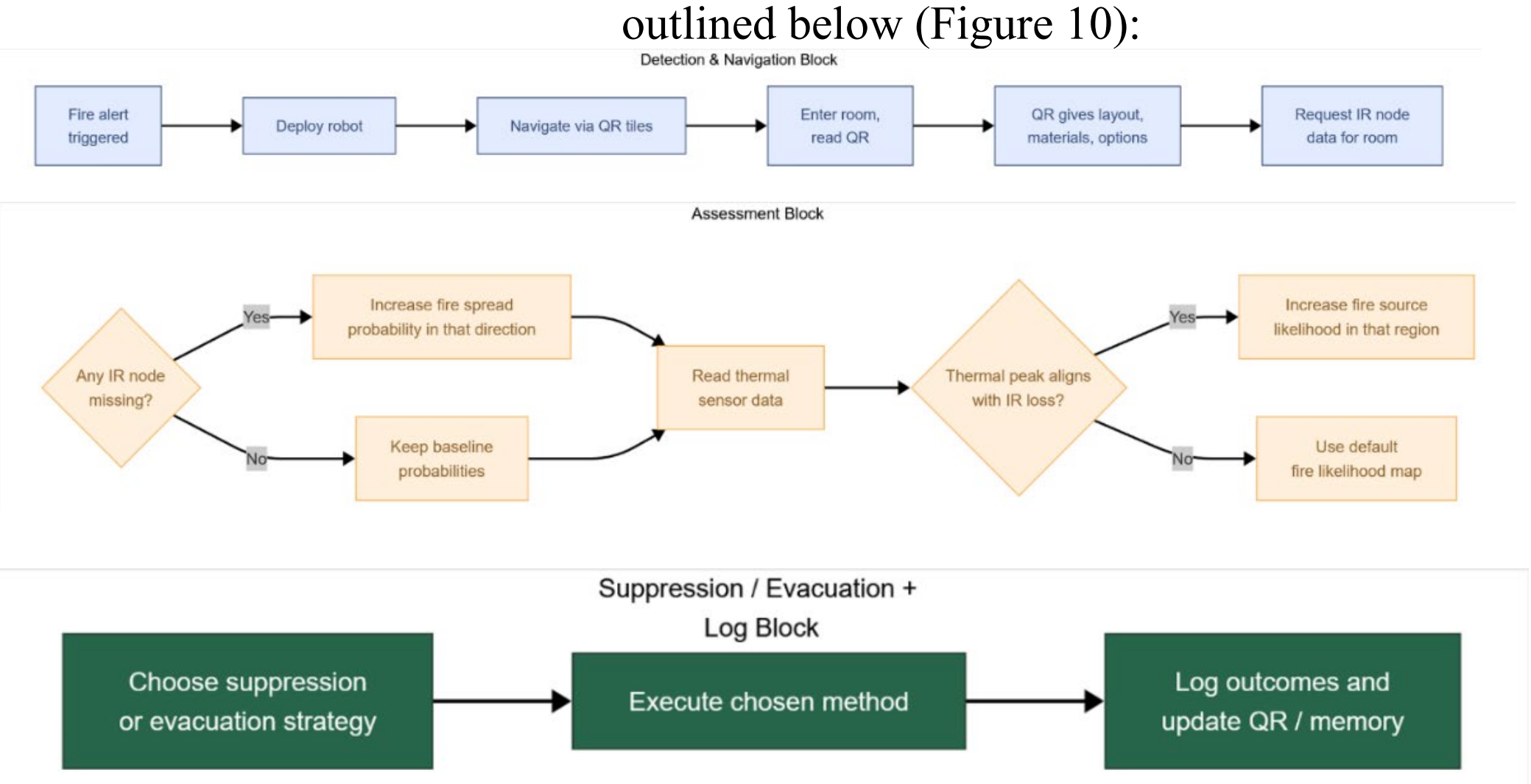

**Figure 10: Details of the Fire-fighting Behavior Architecture and Respective Blocks.**

## SIMULATIONS

To simulate the fire spread, we will be using a cellular automata approach (Was et al., 2020), (Muniyasamy et al., 2024). Due to fire having an indeterminate nature, the basic model will use the basic heat transfer equations for conduction and convection. With the same reasoning, a square grid will be used. Figure 11 shows the cellular automata fire-spread model.

| $\frac{drF(x+1,y+1)}{2}$ | $drF(x,y+1)$ | $\frac{drF(x-1,y+1)}{2}$ |
|---|---|---|
| $drF(x+1,y)$ | $F(x,y)$ | $drF(x-1,y)$ |
| $\frac{drF(x+1,y-1)}{2}$ | $drF(x,y-1)$ | $\frac{drF(x-1,y-1)}{2}$ |

**Figure 11: The Fire spread model describes fire spread from one location to another due to fire flammability risk (r), and the d spread factor, which is dependent on the fire intensity in terms of watts heat. An inflammable material would have a flammability risk of zero.**

While this is definitely not the most true-to-life simulation, the heat transfer equations have been used for an approach like this before and should be accurate enough for the purposes of this paper. The fire-spread cellular automata is being used to simplify fire-spread simulation and make quick responsive decisions. It is based on the current simulation standard developed by NIST and is NIST's Fire Dynamics Simulator (FDS) (McGrattan et al., 2013).

The cellular automata model of the geochemistry lab from Figure 4 is used to perform a first-order comparison between firefighting strategies. Take, for example, the scenario in Figures 12 and 13. In one, the conventional scenario is that a sizeable visible flame is detected before firefighting starts. In our proposed strategy, a sensitive thermal camera detects temperature anomalies and subtle changes, triggering an immediate response.

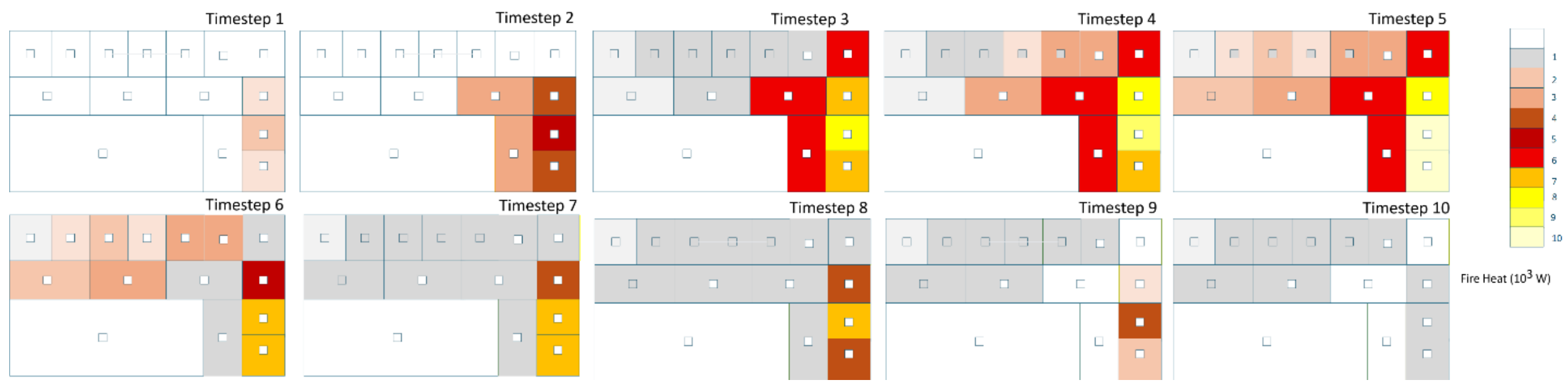

**Figure 12: Conventional fire-fighting response. Fire-fighting actively start at timestep 6 when a significant flame is detected. The first step is to control and surround the fire source. The resultant fire has spread causing extensive damage to facilities.**

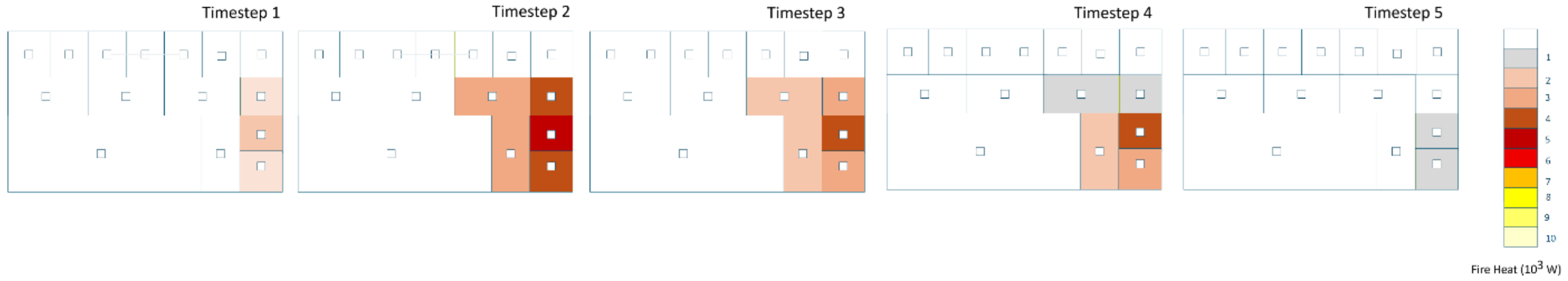

**Figure 13: Proposed Rapid Response to Fire Fighting. Rapid rise in temperatures after timestep 1 triggers response starting at timestep 2. Fire is completely put out by timestep 5, with little trace of damage to facilities.**

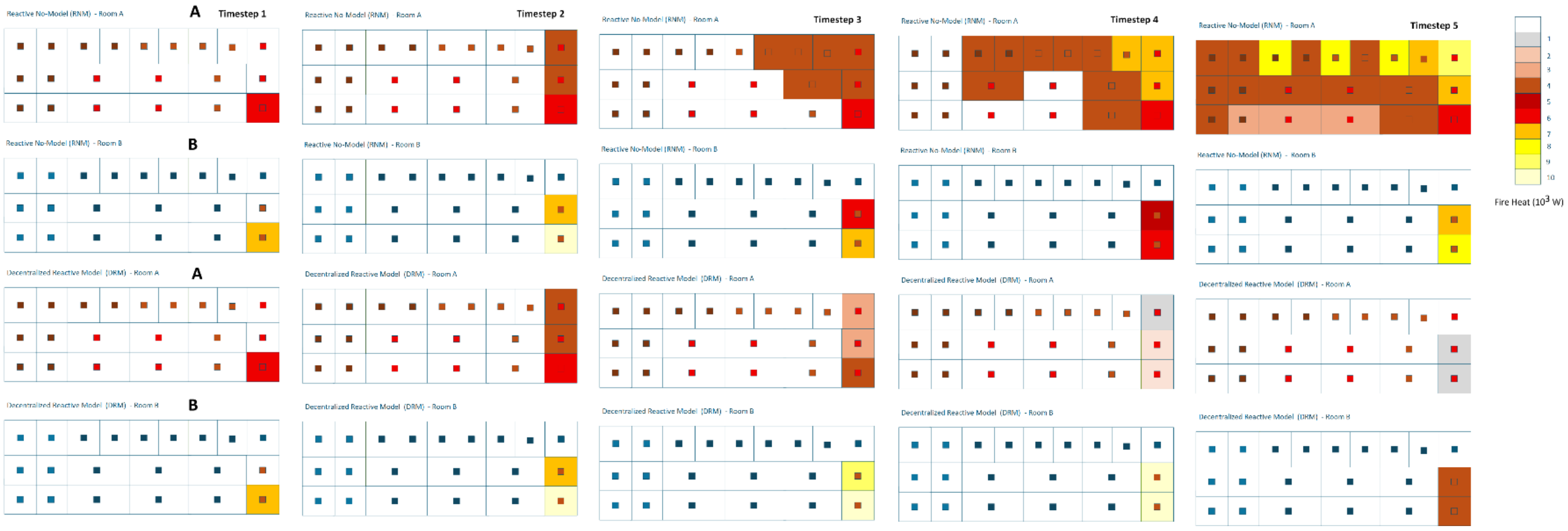

**Figure 14: Room A and Room B are on fire in parallel. In this scenario, we compare conventional no-model reactive response (two right columns) and our proposed model-based approach (two-left columns). The response lets the fire proceed in one room where the chance of spread is zero and works to stop a slow-starting fire, where, unhindered, the fire could turn catastrophic.**

In a second series of scenarios (Figure 14), we compare a conventional reactive fire-fighting approach with our model-based approach. In it, we see two fires co-occurring. One fire happens in Room A. It is a slow-burning fire, but if the fire spreads, it would be catastrophic, as the rest of the room is filled with fire hazards. In contrast, Room B has a higher intensity fire, but the chance of the fire spreading and causing wider damage is next to zero. A conventional reactive response would divert resources to put out the high-intensity fire in Room B. However, this leaves insufficient resources to tackle the more serious (eventual) fire in Room A. Hence, the fire in Room A steadily grows, requiring a conventional response, and eventually becomes overwhelming.

On the other hand, our model-based approach would detect that although Room B has high-intensity fire and the fact that it is burning and well on the way, and further, there is zero risk of spread, it is best not to divert resources and instead work to put out the slow-moving fire in Room A. The strategy results in Room A's fire never being a major problem, and overall results in limited damage to both rooms. Using these simulations, we show that fire-fighting can be a complex decision-making task and benefits from modelling and forecasting to make the optimal decision.

## CONCLUSIONS

In this paper, we present a novel distributed fire-response management system that utilizes QR-Code Tiles embedded in the base to simplify mapping, navigation, and fire-fighting management—the applications of this technology extend beyond lunar habitats to Earth-based facilities. The technology enables the use of simplified, distributed robotic systems to manage firefighting in real time. Early cellular automata simulations demonstrate the approach's efficacy, particularly its rapid, context-aware responses that account for fire-spread risks. Inclusion of these factors in robotic fire-fighting suggests the potential for significant improvements in time, resources, and costs.